\documentclass[letterpaper, 10 pt, conference]{cls/ieeeconf}
\IEEEoverridecommandlockouts
\usepackage{graphics}
\usepackage{graphicx,tabularx,adjustbox}
\usepackage{tikz}
\usetikzlibrary{fadings}
\usetikzlibrary{calc,arrows.meta,decorations.pathreplacing,positioning}
\usepackage{multirow}
\usepackage{booktabs}
\usepackage{algorithm,algorithmicx}
\usepackage[noend]{algpseudocode}
\algnewcommand{\LineComment}[1]{\Statex \(\triangleright\)\ #1}
\usepackage{hhline}
\usepackage{xcolor}
\usepackage{colortbl}
\usepackage{float}
\usepackage{subcaption}
\usepackage{amsmath}
\usepackage{amssymb}
\usepackage{amsfonts,amstext,dsfont,mathtools,bbm}
 
\usepackage{amsthm}
\usepackage{balance}
\usepackage{cite}
\usepackage{xspace}

\makeatletter\let\NAT@parse\undefined\makeatother
\usepackage[bookmarks=true, colorlinks, breaklinks=true]{hyperref}
\usepackage{xurl}

\definecolor{bestcell}{RGB}{198,239,206}
\definecolor{secondcell}{RGB}{255,235,156}
\definecolor{thirdcell}{RGB}{252,213,180}

\newcommand{\floor}[1]{\left\lfloor#1\right\rfloor}

\newcommand{\prl}[1]{\left({#1}\right)}

\newcommand{\crl}[1]{\left\{{#1}\right\}}

\theoremstyle{definition}

\newtheorem*{problem}{Problem}
\theoremstyle{remark}

\newcommand{\bfc}{\mathbf{c}}

\newcommand{\bff}{\mathbf{f}}
\newcommand{\bfg}{\mathbf{g}}

\newcommand{\bfr}{\mathbf{r}}

\newcommand{\bfu}{\mathbf{u}}
\newcommand{\bfv}{\mathbf{v}}

\newcommand{\bfx}{\mathbf{x}}
\newcommand{\bfy}{\mathbf{y}}

\newcommand{\bfphi}{\boldsymbol{\phi}}

\newcommand{\bfxi}{\boldsymbol{\xi}}

\newcommand{\bfD}{\mathbf{D}}

\newcommand{\bfF}{\mathbf{F}}

\newcommand{\bfI}{\mathbf{I}}

\newcommand{\bfW}{\mathbf{W}}

\newcommand{\bfPhi}{\boldsymbol{\Phi}}

\newcommand{\bbR}{\mathbb{R}}
\newcommand{\bbS}{\mathbb{S}}

\newcommand{\calC}{\mathcal{C}}
\newcommand{\calD}{\mathcal{D}}

\newcommand{\calL}{\mathcal{L}}
\newcommand{\calM}{\mathcal{M}}
\newcommand{\calN}{\mathcal{N}}
\newcommand{\calO}{\mathcal{O}}
\newcommand{\calP}{\mathcal{P}}

\newcommand{\calS}{\mathcal{S}}
\newcommand{\calT}{\mathcal{T}}

\newcommand{\calX}{\mathcal{X}}

\def\methodname{OREN-X\xspace}
\def\papertitle{OREN-X: Octree Residual Network for Real-Time Multi-Modal Mapping}
\title{\Large \bf\papertitle}

\author{\authorblockN{Zhirui Dai$^{1}$ \quad Qihao Qian$^{1}$ \quad Dinh Minh Nguyen$^{2}$ \quad Quan-Dung Pham$^{2}$ \quad Kiana Bronder$^{3}$\\ Carlos Nieto-Granda$^{4}$ \quad Yiyu Chen$^{2}$ \quad Quan Nguyen$^{5}$ \quad Nikolay Atanasov$^{1}$}\thanks{$^{1}$The authors are with the Department of Electrical and Computer Engineering, University of California San Diego, La Jolla, CA 92093, USA, e-mails: {\tt\small \{zhdai,\allowbreak q2qian,\allowbreak natanasov\}@\allowbreak ucsd.\allowbreak edu}.}\thanks{$^{2}$Dinh Minh Nguyen, Quan-Dung Pham and Yiyu Chen are with VinMotion, e-mails: {\tt\small v.\allowbreak minhnd59@\allowbreak vinmotion.\allowbreak net, pqdung@\allowbreak snu.\allowbreak ac.\allowbreak kr, yiyuc@\allowbreak usc.\allowbreak edu}.}\thanks{$^{3}$Kiana Bronder is with Parsons, e-mail: {\tt\small kiana.\allowbreak bronder@\allowbreak parsons.\allowbreak us}.}\thanks{$^{4}$Carlos Nieto-Granda is with the U.S. DEVCOM Army Research Laboratory, Adelphi, MD 20783, USA, e-mail: {\tt\small carlos.\allowbreak p.\allowbreak nieto2.\allowbreak civ@\allowbreak army.\allowbreak mil}.}\thanks{$^{5}$Quan Nguyen is with the University of Southern California, Los Angeles, CA 90089, USA, e-mail: {\tt\small quann@\allowbreak usc.\allowbreak edu}.}}

\begin{document}
\bstctlcite{IEEEexample:BSTcontrol}

\maketitle
\thispagestyle{empty}
\pagestyle{empty}

\begin{abstract}

To achieve general-purpose autonomy over long horizons, a robot needs to maintain spatial environment information that supports a variety of tasks: geometry for planning and control, radiance for rendering and relocalization, and vision-language features for open-vocabulary grounding.
Existing methods represent and estimate each modality separately, multiplying memory and compute cost while forgoing potential synergy among the representations.
We develop \methodname, an online mapping method that uses an octree in 3D space as a shared data structure for indexing and storing a multi-modal field, capturing geometric, radiance, and vision-language information. \methodname provides efficient unified storage and retrieval of these data in explicit/implicit and full/compressed form.
Our unified representation yields cross-modality synergy: SDF estimates are sharpened by occupancy and radiance, while GPU-based ray-octree traversal and octree query enable real-time rendering.
We also use online dictionary learning to compress the vision-language features, shrinking them $3.7\times$ below full per-vertex storage while raising the query accuracy.
On Replica, \methodname maps in real time ($80{+}$\,fps for SDF and $30{+}$\,fps for all four modalities), improves near-surface SDF accuracy by $33\%$ over single-modality baselines, and improves mean open-vocabulary 3D mIoU by $71\%$ and mean accuracy by $61\%$ over the best prior method.
\end{abstract}

\section{Introduction}
\label{sec:intro}

Long-horizon general-purpose autonomy requires a robot to maintain multiple types of information about its environment.
Modern robotics stacks often maintain a separate representation per modality, such as a geometric field (distance or occupancy) for planning and control, a radiance field for rendering and relocalization, and a vision-language field for open-vocabulary querying and instruction grounding.
This is inefficient in two ways. First, it multiplies memory and computation with the number of representations, a critical bottleneck for real-time, long-horizon operation. Second, independent modalities cannot exploit synergies among them.

Many existing methods target a single modality. Online SDF mapping fuses truncated SDF into discrete grids \cite{oleynikova_voxblox_2017, millane_nvblox_2024} or fits continuous neural fields \cite{ortiz_isdf_2022, hio-sdf_2024, pan_pin-slam_2024, oren26}. NeRF- and 3DGS-based methods learn radiance \cite{h3mapping2024, gs2023}. A growing body of work learns vision-language (VL) fields \cite{lerf2023, langsplat2024, latentam2026}.
A robot that needs all of these capabilities must run several pipelines side by side.
Storing multiple modalities in one spatial structure is not enough to unify them.
Unless the modalities also share how their parameters are stored, queried, and trained, each keeps its own parameters and updates, benefiting from none of the others.
A unified representation therefore needs three properties: common storage across modalities, one indexing scheme that queries them together, and joint training in which each modality improves the others.

\begin{figure}
    \centering
    \includegraphics[width=\linewidth]{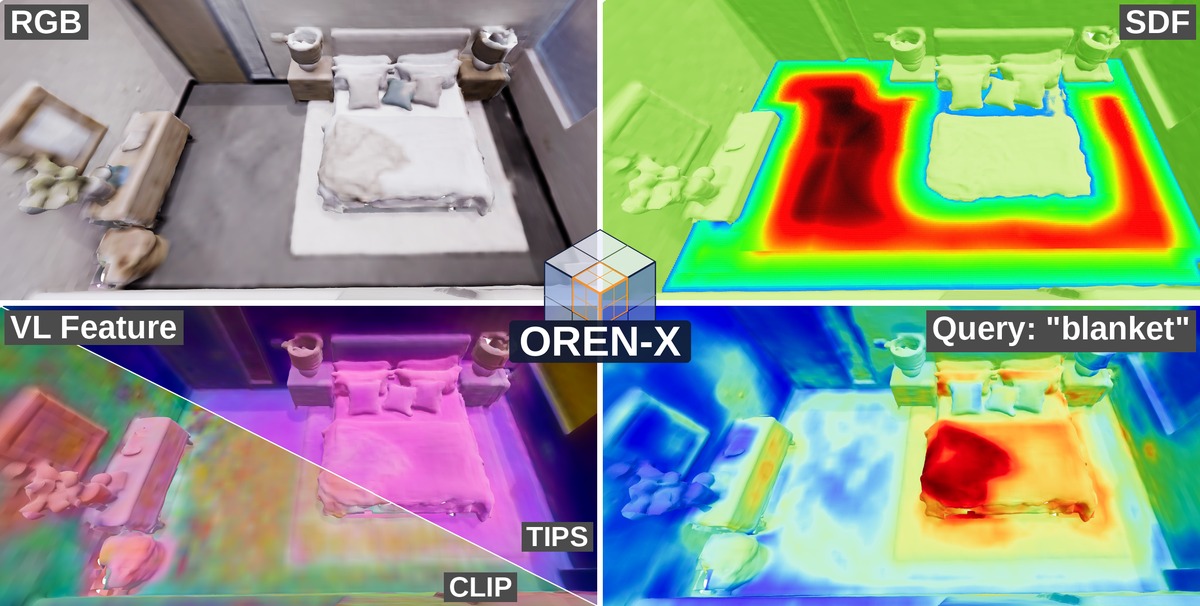}
    \caption{\textbf{\methodname at a glance.} From RGB-D data, \methodname builds a single octree-based multi-modal field that jointly represents geometry, radiance and vision-language features (shown here for two backbones, CLIP and TIPS), supporting real-time open-vocabulary queries (e.g.\ ``blanket'') localized directly in 3D.}
    \label{fig:teaser}
\end{figure}

We present \methodname, an online multi-modal mapping framework that represents all modalities with a single octree.
Every modality stores its parameters, as explicit values, latent features decoded by a small network, or both, at the octree vertices, in full or compressed form to save memory.
All modalities use the same octree lookup to locate and interpolate their parameters to obtain continuous values at any query point, and are optimized together online with per-modality and cross-modality losses that let one modality improve another.
We demonstrate \methodname with four modalities: signed distance, occupancy, radiance and vision-language.

Joint training yields cross-modality synergy.
Radiance supervision through an SDF-derived density improves SDF accuracy, and an SDF-occupancy consistency loss improves robustness under strong depth noise.
The shared octree makes the VL features queryable in 3D by interpolation and in 2D by rendering with radiance.
The octree also reduces computation because it supports both GPU ray-octree traversal for rendering and fast tile-based rasterization.
Because the VL features need the most memory, we develop an online dictionary learning method that compresses them, improves query accuracy, and scores text queries directly in low-dimensional code space.
Our contributions are as follows.
\begin{itemize}
    \item \textbf{Unified multi-modal mapping.} We develop an octree that provides common storage, shared indexing, and joint training for different modalities, such as signed distance, occupancy, radiance, and vision-language.

    \item \textbf{Cross-modality synergy.} Radiance supervision improves SDF accuracy, an SDF-occupancy consistency loss improves noise robustness, and shared indexing enables both 2D and 3D open-vocabulary queries.

    \item \textbf{Memory and compute efficiency.} Shared indexing lets one octree traversal serve all modalities in real time. We develop an online dictionary learning method to compress the VL features with higher query accuracy and to answer text queries directly in code space.
\end{itemize}

\section{Related Work}
\label{sec:related}

\noindent\textbf{Online geometric mapping.}
Dense geometric mapping has long relied on volumetric fusion of depth into discrete grids, from early truncated-SDF fusion \cite{curless_volumetric_1996} made real-time on a GPU by KinectFusion \cite{newcombe_kinectfusion_2011}, to probabilistic occupancy octrees that bound memory and distinguish free from unknown space for planning \cite{hornung_octomap_2013}, to incremental Euclidean SDF extraction from the TSDF for collision checking on CPU \cite{oleynikova_voxblox_2017} and GPU \cite{millane_nvblox_2024}.
These grids, however, are limited by voxel resolution. Occupancy Networks instead represent geometry as a continuous neural decision boundary \cite{mescheder_occupancy_2019}.
Recent methods fit continuous, neural or hybrid SDF fields online \cite{ortiz_isdf_2022, hio-sdf_2024}, often on hierarchical feature grids that concentrate parameters near surfaces \cite{takikawa_lod_2021, instant-ngp, h3mapping2024}.
An octree suits such sparse 3D data, offering fast lookup, multi-resolution detail, and incremental growth \cite{takikawa_lod_2021, svraster2025}. OREN \cite{oren26} achieves online SDF reconstruction using an explicit coarse prior and an implicit neural residual. A single geometric field, however, does not provide semantic information a robot needs to execute complex tasks autonomously.

\noindent\textbf{Radiance fields and rendering.}
Neural radiance fields (NeRF) \cite{nerf2020} and their accelerated grid variants \cite{instant-ngp, plenoctrees2021} render appearance by volumetric ray marching, while surface-oriented formulations tie appearance to geometry through an SDF-based density term \cite{wang_neus_2021}.
3D Gaussian Splatting (3DGS) \cite{gs2023} and its surface-accurate variants \cite{gs2d2024} instead render appearance by fast tile-based rasterization, and sparse-voxel rasterizers reach comparable quality with voxel primitives \cite{svraster2025}.
These two families, ray-based and tile-based rasterization, trade query flexibility against speed, and our method supports both on one octree, where GPU-based ray-octree traversal and query keep the ray-based rasterizer efficient for real-time radiance optimization while the tile-based rasterizer renders images faster.

\noindent\textbf{Open-vocabulary language fields.}
Semantic octrees \cite{asgharivaskasi_semantic_2023} and 3D scene graphs such as Hydra \cite{hughes_hydra_2022} attach semantic labels to geometry but their label sets are fixed before deployment.
Vision-language (VL) backbones such as CLIP \cite{clip2021}, DINOv3 \cite{dinov3} with text alignment \cite{dinotxt2024, talk2dino2025}, and TIPSv2 \cite{tipsv2_2026} remove this restriction by embedding images and text in a shared feature space, but since these features live in the image plane, they must be fused into a spatial representation to become consistent across views and queryable in 3D.
LERF distills VL features into a NeRF \cite{lerf2023}, 3DGS-based methods store them in each Gaussian \cite{langsplat2024, langsplatv2_2025, legaussians2026}, and point- or graph-based maps attach them to points or objects \cite{openscene2022, hovsg2024, octreegraph2024}, usually compressing the memory-prohibitive per-primitive features by autoencoders \cite{langsplat2024, onlinelangsplat2025}, quantization \cite{legaussians2026, langsplatv2_2025}, or online dictionary learning \cite{mairal2009odl, latentam2026}.
These methods, however, layer the language field on an appearance reconstruction or keep it as a separate semantic map rather than maintaining it with the geometry a robot needs for navigation.
In contrast, \methodname maintains VL features jointly with geometry and appearance in one octree, and improves memory and open-vocabulary retrieval using online dictionary learning of a compressed VL feature code.

\section{Problem Statement}
\label{sec:problem}

We aim to reconstruct a single multi-modal field $f = (f_c)_{c \in \calM}$ with four modalities, $\calM = \crl{\mathrm{sdf}, \mathrm{occ}, \mathrm{rad}, \mathrm{vl}}$: signed distance ($\mathrm{sdf}$), occupancy ($\mathrm{occ}$), radiance ($\mathrm{rad}$) and vision-language features ($\mathrm{vl}$).
The component $f_c : \calX_c \to \bbR^{d_c}$ of modality $c$ has domain $\calX_c = \bbR^3$ for $\mathrm{sdf}$, $\mathrm{occ}$ and $\mathrm{vl}$, and $\calX_\mathrm{rad} = \bbR^3 \times \bbS^2$, which adds the viewing direction.

At time $t$, a robot receives RGB image $\bfI_t$, depth image $\bfD_t$ and camera pose $\bfxi_t \in SE(3)$, and a VL model computes pixel-aligned features $\bfPhi_t$.
Back-projecting pixels with depth and pose gives training data $\calD_{c,t} = \crl{(\bfx_i, \bfy_i)}_i$ per modality $c$, where $\bfx_i$ is a 3D position or camera ray and $\bfy_i \in \bbR^{d_c}$ is its label.
The RGB image provides pixel colors for $\mathrm{rad}$, and the depth image provides observed surface points for $\mathrm{sdf}$ and $\mathrm{occ}$.
For $\mathrm{sdf}$ and $\mathrm{occ}$, we sample points along each ray, in free space before the surface point and around it.
The $\mathrm{sdf}$ label of a sample is its distance to the nearest observed surface, negated behind it, and the $\mathrm{occ}$ label is $0$ in free space, $0.75$ at the surface (logit $\approx 1.0$), and $1$ behind it.

\begin{problem}
Given the stream of RGB images $\bfI_\tau$, depth images $\bfD_\tau$, camera poses $\bfxi_\tau$ and vision-language features $\bfPhi_\tau$ for $\tau \le t$, estimate online the multi-modal field $f$ that minimizes
\begin{equation}
    \sum_{j} \lambda_j\, \mathcal{L}_j\!\prl{\crl{f_c}_{c \in \calS_j};\ \calD_t},
    \label{eq:problem}
\end{equation}
with bounded computation and memory, where $\calD_t = \bigcup_{c,\,\tau \le t} \calD_{c,\tau}$, and each loss term $\mathcal{L}_j$ with weight $\lambda_j$ depends on a subset of modalities $\calS_j \subseteq \calM$ and evaluates them at positions or integrates them along rays.
A term with $|\calS_j| > 1$ is a cross-modality loss that couples the modalities in $\calS_j$.
\end{problem}

\section{Technical Approach}
\label{sec:approach}

\methodname performs online multi-modal mapping: from a posed RGB-D stream, it estimates all four modalities of one field in real time with a single octree.
Sec.~\ref{sec:unified_map} develops an octree representation of the multi-modal field and its online update.
Sec.~\ref{sec:synergy} describes how each modality is decoded and supervised, including the cross-modality loss terms.
Sec.~\ref{sec:vl} presents our online dictionary learning algorithm that compresses the VL features.

\subsection{Unified Multi-Modal Mapping}
\label{sec:unified_map}

Except for learnable weights $\bfW$ shared across vertices by the decoder $g_c$, \methodname stores all modalities' parameters at the octree vertices, organized by three designs.
First, multi-pool indexing lets each modality share the octree at its own spatial resolution.
Second, all modalities share one per-vertex parameterization, where a vertex stores explicit values, implicit features, or both, in full or compressed form, so each modality uses the combination that suits it.
Third, an active-vertex online update keeps the per-frame cost bounded.

\begin{figure}[t]
    \centering
    \resizebox{\linewidth}{!}{\input{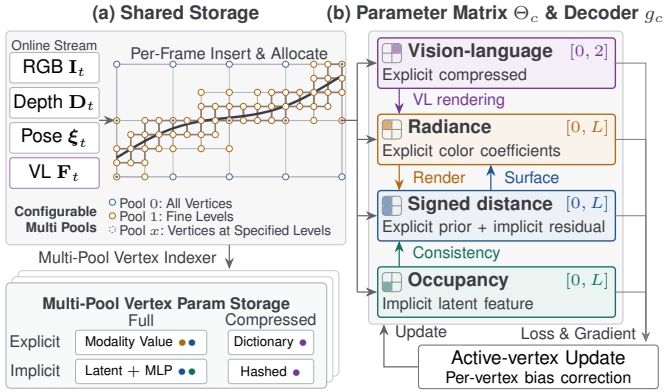}}
    \caption{\textbf{\methodname overview.}
        (a) A stream of RGB-D and VL features updates a multi-modal field with four modalities ($\mathrm{sdf}$, $\mathrm{occ}$, $\mathrm{rad}$, and $\mathrm{vl}$). Observed points allocate octree voxels at different levels, and multi-pool indexing maps a query point to its surrounding vertices' rows in each modality's parameter matrix $\Theta_c$.
        (b)~Each modality's parameters are explicit values, implicit features, or both, stored in full or compressed form, decoded by the decoder $g_c$ to produce modality values and loss terms. The loss terms are summed and optimized by a single backward pass, updating only active vertices with one Adam step using per-vertex bias correction.}
    \label{fig:overview}
\end{figure}

\noindent\textbf{Octree parameterization.}
\methodname represents the field $f$ on an octree $\calO$, which partitions space into cubic voxels at levels $\ell = 0, \dots, L$, where level $0$ is the finest with size $r_0$ and level $\ell$ has size $r_\ell = r_0 2^\ell$.
Each allocated voxel's parent is also allocated, so voxels form a tree, with each voxel's eight corners as its vertices.
Each modality $c \in \calM$ has a parameter matrix $\Theta_c \in \bbR^{V_c \times n_c}$, whose row $\Theta_c[k]$ is the parameter vector of the $k$-th vertex allocated to $c$.
An indexing function $\calP_c$ maps a query position $\bfx$ to the vertices around it, and a decoder $g_c$, with weights $\bfW$, predicts modality $c$'s value,
\begin{equation}
    f_c(\bfu) = g_c\!\prl{\bfu,\ \textstyle\sum_{k \in \calP_c(\bfx)} w_k(\bfx)\, \Theta_c[k]} ,
    \label{eq:mapstructure}
\end{equation}
where $\bfu \in \calX_c$ is the input, $\bfx$ is its position part, and $w_k(\bfx)$ are interpolation weights, e.g., trilinear.
\methodname estimates $\crl{\Theta_c}_{c \in \calM}$ and $\bfW$ online to optimize \eqref{eq:problem}, as shown in Fig.~\ref{fig:overview}.

\noindent\textbf{Multi-pool indexing.}
\methodname gives each modality $c \in \calM$ its own vertex pool spanning a contiguous level range $\calL_c = \crl{\ell : \ell_c^\mathrm{lb} \le \ell \le \ell_c^\mathrm{ub}}$, so heterogeneous modalities share one octree at different spatial resolutions and extents.
Each vertex $\bfv$ is keyed by its Morton code \cite{svraster2025},
\begin{equation}
    \calT(\bfv, r_0) = \mathrm{Morton}(\floor{\bfv/r_0}), \label{eq:morton}
\end{equation}
which interleaves the bits of the three integer coordinates $\floor{\bfv/r_0}$ into a single, level-independent integer.
Using this key, the indexing function $\calP_c$ gathers, for every $\ell \in \calL_c$, the eight corners $\calC_\ell(\bfx)$ of the level-$\ell$ voxel containing $\bfx$,
\begin{equation}
    \calP_c(\bfx) = \crl{(\ell, \calT(\bfv, r_0)) : \ell \in \calL_c,\
    \bfv \in \calC_\ell(\bfx)} .
    \label{eq:index}
\end{equation}
This multi-pool design allows each modality to pick its own level range and allocate parameters only where needed, while still sharing the same octree for efficient indexing and update.

\noindent\textbf{Per-vertex representation.}
Each row $\Theta_c[k]$ can contain explicit values and implicit features.
The explicit part stores the modality's value or spatial derivatives at the vertex, such as an SDF value with its gradient or a color, so interpolating it directly estimates $f_c$.
The implicit part stores a latent feature that $g_c$ decodes into the modality value or a residual correcting the explicit part's estimate of $f_c$.
Either part can instead be stored in compressed form, as a shorter vector of weights over a shared fixed-capacity dictionary, from which the full vector is recovered as a weighted sum of dictionary atoms.
Fig.~\ref{fig:overview} shows each modality's configuration, chosen by role and empirical experience.
SDF keeps OREN's explicit prior \cite{oren26} for coarse geometry and adds an implicit residual for detail. Occupancy decodes the implicit feature with its own MLP. Radiance stores colors explicitly for direct rendering. VL features are stored explicitly for linear decoding, so text queries score directly in the code space (Sec.~\ref{sec:vl}), and are compressed by our online dictionary learning algorithm.
This parameterization admits other combinations of explicit, implicit, and compressed representations, but ablating them is beyond the scope of this paper.

\noindent\textbf{Online update.}
Following OREN \cite{oren26}, our formulation grows the octree with each new frame and maintains a key-frame set to sample training frames alongside the current one.
Streaming, partially observed supervision updates each vertex a different number of times, since vertex $k$ gets a gradient only on frames where it is active, so its effective step count $t_k$ varies across vertices and departs from the global iteration counter.
Vanilla Adam, whose bias correction assumes every parameter is stepped every iteration, over-shrinks moment estimates for rarely observed vertices. We instead keep a per-vertex step count $t_k$, incremented only when $k$ is active, to bias-correct the first- and second-moment estimates:
\begin{equation*}
    m_k \leftarrow \beta_1 m_k + (1-\beta_1)\, g_k, \quad
    v_k \leftarrow \beta_2 v_k + (1-\beta_2)\, g_k^2 ,
\end{equation*}
\begin{equation}\label{eq:onlineadam}
    \Theta_c[k] \leftarrow \Theta_c[k] - \alpha\,
    \frac{m_k / (1-\beta_1^{t_k})}{\sqrt{v_k / (1-\beta_2^{t_k})} + \epsilon},
\end{equation}
where $\alpha,\beta_1,\beta_2,\epsilon$ are the usual Adam constants and $g_k$ is the accumulated gradient for $\Theta_c[k]$.
This per-vertex update is also sparse, since only active vertices carry non-zero gradients, so the moment and parameter updates touch those vertices and per-frame cost scales with the observed region rather than the global map. The active set is the union of all modalities' active vertices, recorded by octree queries.

\subsection{Cross-Modality Synergy}
\label{sec:synergy}

Co-locating modalities on one octree lets them reinforce one another in three ways:
(i) \emph{quality}: modalities supervise one another, (ii) \emph{compute}: one top-down octree descent, ray-octree traversal, or in-frustum voxel filtering serves several modalities at once, (iii) \emph{memory}: the shared allocation, indexing and updates are paid once, rather than $|\calM|$ times, across modalities. We describe each modality below and emphasize its coupling to the multi-modal representation.

\noindent\textbf{Signed distance.}
We adopt OREN's SDF representation and loss terms \cite{oren26}.
Each vertex $k$ with position $\bfx_k \in \bbR^3$ stores an SDF value $d_k \in \bbR$ and gradient $\bfg_k \in \bbR^3$ as the explicit part, and a feature $\bff_k \in \bbR^{16}$ as the implicit part, so $\Theta_\mathrm{sdf}[k] = (d_k, \bfg_k, \bff_k)$.
Following \eqref{eq:mapstructure}, the decoder $g_\mathrm{sdf}$ extrapolates the explicit part of the finest voxel's corner $k \in \calP_\mathrm{sdf}(\bfx)$ to the query point, $d_k(\bfx) = d_k + \bfg_k^\top(\bfx - \bfx_k)$, and interpolates the results with the trilinear weights $w_k(\bfx)$ into a coarse estimate $d_\mathrm{ga}(\bfx)$.
A small MLP $D$ with weights $\bfW$ then adds a residual from $d_\mathrm{ga}(\bfx)$ and the interpolated feature $\bff = \sum_k w_k(\bfx)\,\bff_k$, $f_\mathrm{sdf}(\bfx) = d_\mathrm{ga}(\bfx) + D(d_\mathrm{ga}(\bfx), \bff; \bfW)$.
Since a valid SDF has unit gradient norm, we regularize the gradient with the Eikonal loss, $\big|\lVert \nabla f_\mathrm{sdf}(\bfx) \rVert - 1\big|$, computed by finite differences. To speed it up, a CUDA kernel looks up the sample point's containing voxel and its six stencil points simultaneously, since paired stencil points are usually in the same voxel. The kernel executes an extra descent only when a stencil point falls into a neighboring voxel, so we reuse shared indexing without an extra octree traversal (\emph{compute}).

\noindent\textbf{Occupancy.}
The occupancy at a point $\bfx$ is predicted as a logit $\hat{\ell}(\bfx)$, decoded by an MLP from its own implicit feature, and supervised by the occupancy label $y(\bfx)$ with a binary cross-entropy (BCE) loss.
A sign-consistency term then lets the detached occupancy teach the SDF its sign, penalizing SDF predictions $\hat{s}(\bfx)$ that fall on the wrong side of a confident occupancy logit,
\begin{equation}
\begin{aligned}
    \mathcal{L}_{\mathrm{occ}}  &= \operatorname{BCE}\!\big(\hat{\ell}(\bfx),\, y(\bfx)\big), \\
    \mathcal{L}_{\mathrm{cons}} &= \max\!\big(0,\ \hat{s}(\bfx)\,\operatorname{sg}[\operatorname{sign}\hat{\ell}(\bfx)]\big) ,
\end{aligned}
    \label{eq:occ}
\end{equation}
with $\mathcal{L}_{\mathrm{cons}}$ restricted to confident points $|\hat{\ell}(\bfx)| > \eta$, and $\operatorname{sg}[\cdot]$ the stop-gradient. Because $\hat{\ell}$ is detached, occupancy makes the SDF sign robust under strong depth noise without the SDF pulling occupancy back (\emph{quality}).

\noindent\textbf{Radiance.}
The radiance modality stores an RGB color at each vertex, so $n_\mathrm{rad} = 3$, supervised by images.
Each pixel defines a ray $\bfr$ through it from the camera center, and we predict its color $\hat{C}(\bfr)$ from the field.
Following volume rendering \cite{nerf2020}, we evaluate the field at $K$ points along the ray from the camera to the surface, and composite their colors,
\begin{equation}
    \hat{C}(\bfr) = \sum_{k=1}^{K} T_k\,\alpha_k\,\bfc_k, \quad
    T_k = \prod_{j<k}(1-\alpha_j),
    \label{eq:volrender}
\end{equation}
where $\bfc_k$ is the color decoded at sample $k$, $\alpha_k = 1 - e^{-\sigma_k \delta_k}$ is the probability the ray stops between sample $k$ and the next one at distance $\delta_k$, and $T_k$ is the probability the ray reaches sample $k$.
Instead of learning a separate density $\sigma_k$, we derive it from the SDF \cite{wang_neus_2021}, so the photometric loss on $\hat{C}(\bfr)$ couples the SDF and radiance modalities.
The SDF-induced surface location lets the renderer sample on the surface more accurately, and photometric gradients flow back through the density term to refine the SDF, so SDF and radiance improve each other (\emph{quality}).
We adopt the tile-based rasterizer of \cite{svraster2025} for real-time octree-based rendering, but since it only touches in-frustum voxels per frame and converges slowly during training, we add a ray-based rasterizer, whose GPU kernel implements the parametric ray-octree traversal \cite{octreetraversal2000}, that trains more efficiently and converges faster. We thus train with the ray-based rasterizer and render with the tile-based one on the same octree (\emph{compute}).

\noindent\textbf{Vision-language.}
We store the VL features in a near-surface band of the octree, co-located with the radiance modality. So, \methodname supports both rendering the color and the VL features within the same 2D rendering pass, as well as querying the VL features in 3D space (\emph{compute}). It is prohibitive to store the VL features in full dimension for every vertex, so we compress them into a sparse code over a shared dictionary (\emph{memory}), detailed in Sec.~\ref{sec:vl}.

\subsection{Memory Compression for Vision-Language Features}
\label{sec:vl}

Unlike most existing methods that require pre-trained dictionaries \cite{langsplatv2_2025,legaussians2026}, \methodname learns the dictionary online from the streaming features.

\noindent\textbf{Online dictionary.}
We use an append-only rule to grow the basis elements (atoms) of a dictionary. When current atoms leave enough energy of new observations unexplained, a new atom is added and never modified.
The dictionary resembles online PCA \cite{onlinepca2018} for VL features, except its atom count grows by an energy criterion and the atoms do not rotate with the subspace.
Each vertex stores a vector of weights (code) over the atoms, instead of the raw feature.
As a vertex is re-observed and the dictionary grows, its code is updated with the atoms fixed. Alg.~\ref{alg:codebook} summarizes the procedure.

\begin{algorithm}[t]
\caption{Online dictionary: accumulate, initialize, grow.}
\label{alg:codebook}
\small
\begin{algorithmic}[1]
\Require decay $\rho$, energy threshold $\epsilon_{\mathrm{dict}}$, capacity $G_{\max}$
\State $D \gets [\,]$, $R \gets \mathbf{0}$, $E \gets 0$ \Comment{dictionary, residual scatter, energy}
\Procedure{Accumulate}{$V$}\Comment{$V \in \mathbb{R}^{M\times d}$, one frame}
    \State $A \gets V D$;\quad $R_\perp \gets V - A D^\top$
        \LineComment{projected codes and orthogonal-complement residual}
    \State $R \gets \rho R + R_\perp^\top R_\perp$;\quad
            $E \gets \rho E + \lVert V\rVert_F^2$
\EndProcedure
\Procedure{Bootstrap}{}\Comment{one-shot initialization}
    \State $(\Lambda, U) \gets \textsc{eigh}(R)$
    \State $D \gets [\,u_j : \lambda_j / E > \epsilon_{\mathrm{dict}}\,]$
            \Comment{keep dominant directions}
    \State $R \gets R - \textstyle\sum_j \lambda_j\, u_j u_j^\top$
            \Comment{deflate absorbed energy}
\EndProcedure
\Procedure{Grow}{}\Comment{append-only}
    \While{$|D| < G_{\max}$}
        \State $(\lambda, u) \gets \Call{PowerIter}{R, D}$
        \If{$u = \varnothing$ \textbf{ or } $\lambda / E \le \epsilon_{\mathrm{dict}}$}
            \State \textbf{break}
        \EndIf
        \State $D \gets [\,D,\, u\,]$;\quad $R \gets R - \lambda\, u u^\top$
    \EndWhile
\EndProcedure
\Function{PowerIter}{$C, D$}
    \State $\textsc{Proj}(x) \triangleq x - D D^\top x$
    \If{$|D| \ge d$}\Comment{complement empty} \State \Return $(0, \varnothing)$ \EndIf
    \State $u \gets \textsc{Proj}(x)$ for a random unit $x$;\quad $u \gets u / \lVert u\rVert$
    \Repeat
        \State $v \gets \textsc{Proj}(C u)$;\quad $v \gets v / \lVert v\rVert$
        \State $\Delta \gets \lVert v - u\rVert$;\quad $u \gets v$
    \Until{$\Delta < \delta$ \textbf{ or } max iterations}
    \State \Return $(u^\top C u,\ u)$\Comment{Rayleigh quotient, unit eigenvector}
\EndFunction
\end{algorithmic}
\end{algorithm}

\noindent\textbf{Text query.}
A text query is embedded by a text encoder as $\bfphi_q$ and scored against a VL feature $\bff$ by cosine similarity,
\begin{equation}
    \mathrm{sim}(\bff, \bfphi_q) = \frac{\bff^\top \bfphi_q}
    {\lVert \bff \rVert\, \lVert \bfphi_q \rVert}.
    \label{eq:cossim}
\end{equation}
Following \cite{lerf2023}, a relevancy score contrasts the query with a set of negative queries $\crl{\bfphi_n^i}$ and keeps the least favorable,
\begin{equation}
    \mathrm{rel}(\bff, \bfphi_q) = \min_i
    \frac{e^{\mathrm{sim}(\bff,\bfphi_q)/\tau}}
    {e^{\mathrm{sim}(\bff,\bfphi_q)/\tau} + e^{\mathrm{sim}(\bff,\bfphi_n^i)/\tau}} ,
    \label{eq:relevancy}
\end{equation}
where $\tau$ scales the similarities.
Because decoding is linear, a text query is embedded once into the code space $\bbR^{|D|}$ and scored directly against per-vertex codes in the dictionary's coordinates, without decoding back to the $d_{\mathrm{vl}}$ feature.

\begin{figure}[t]
    \centering
    \includegraphics[width=\linewidth]{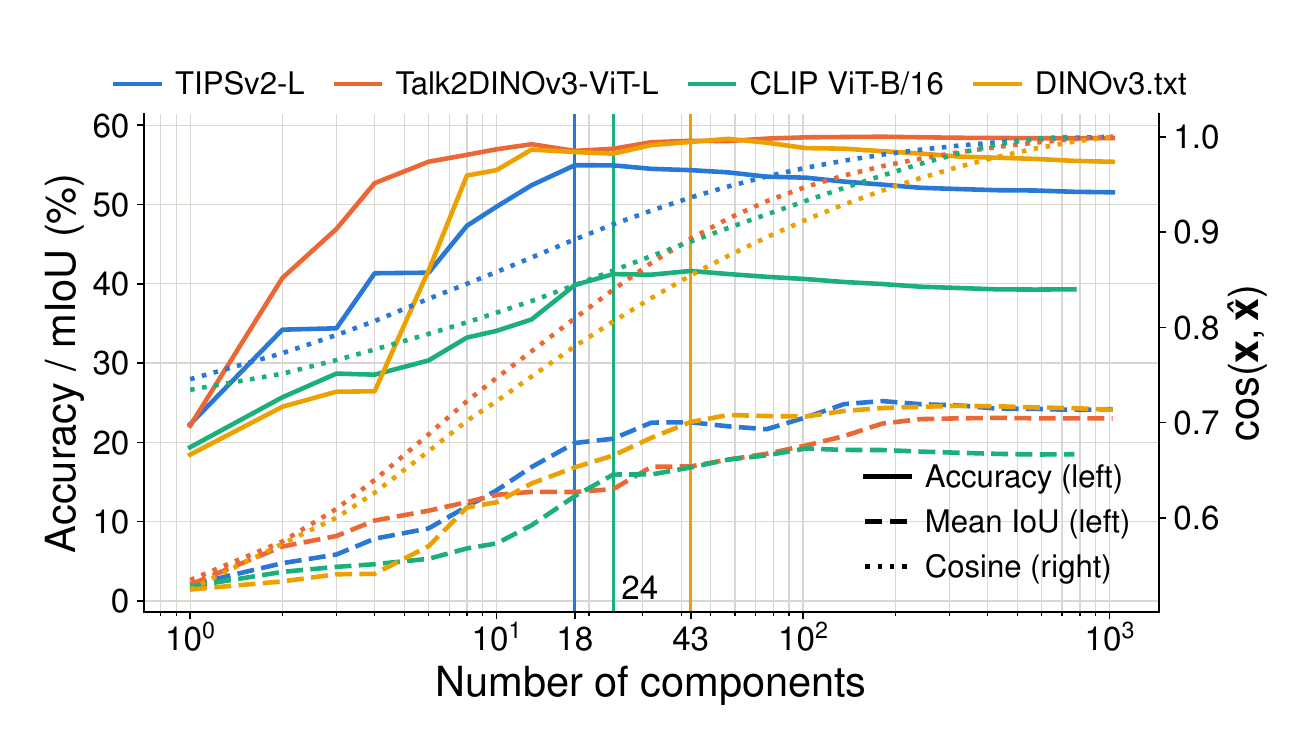}
    \caption{\textbf{VL reconstruction vs.\ query accuracy.} As the dictionary size grows, open-vocabulary query accuracy saturates well before reconstruction fidelity does.}
    \label{fig:cb_saturate}
\end{figure}

\begin{figure}[t]
    \centering
    \includegraphics[width=\linewidth]{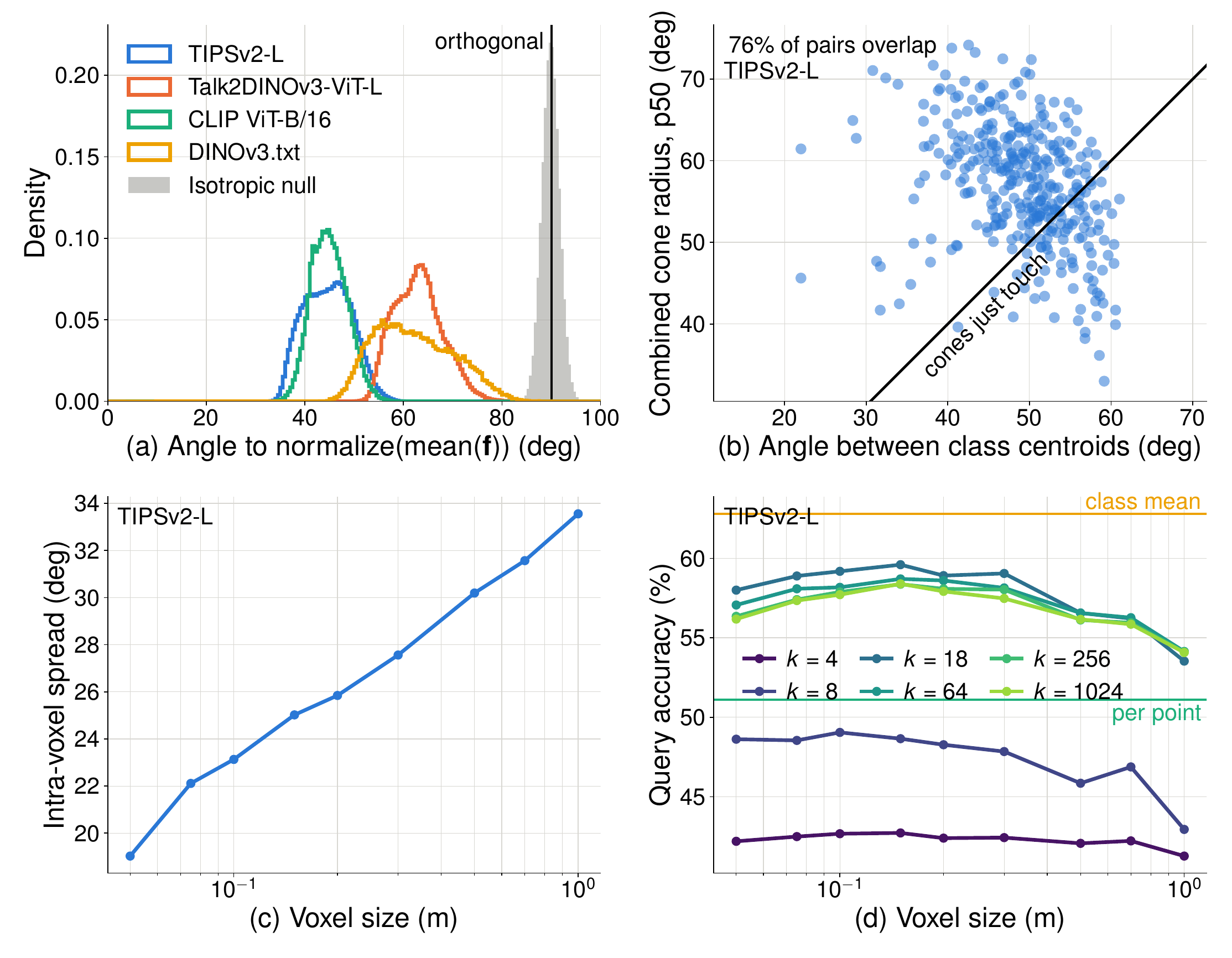}
    \caption{\textbf{Geometry of VL features in Replica \cite{replica19arxiv} \texttt{room0}.}
        (a)~Histogram of angles between each feature and the normalized mean feature. Features lie in a $35$-$80^\circ$ cone around the mean, far from the $90^\circ$ expected for isotropic features.
        (b)~Combined median cone radius vs.\ the angle between class centroids, per class pair. $76\%$ of pairs overlap (above the diagonal), so a single feature is often ambiguous between classes.
        (c)~Feature angular spread within a voxel grows with voxel size.
        (d)~Query accuracy of per-voxel PCA features ($k$ components) vs.\ voxel size, between per-point and class-mean references. Accuracy saturates by $k{=}18$ and peaks at intermediate voxel sizes.} \label{fig:vl_geometry}
\end{figure}

\noindent\textbf{Compression analysis.}
Open-vocabulary querying scores features by cosine similarity \eqref{eq:cossim} and relevancy \eqref{eq:relevancy}, which depend on the few directions that separate concepts, not on faithful reconstruction.
As shown in Fig.~\ref{fig:cb_saturate}, the reconstruction fidelity is affected by low-variance directions that leave query-score ranking unchanged. As the dictionary grows, query accuracy saturates while reconstruction fidelity keeps improving, so the dictionary should only admit directions up to the saturation point.
As shown in Fig.~\ref{fig:vl_geometry}a, VL features typically concentrate in a cone around their mean rather than spreading isotropically, so a few directions capture most of their energy.
The cones of most class pairs overlap (Fig.~\ref{fig:vl_geometry}b), so a single feature is often ambiguous between classes, and aggregating features within a voxel raises query accuracy above the per-point level (Fig.~\ref{fig:vl_geometry}d).
Although feature spread within a voxel grows with voxel size (Fig.~\ref{fig:vl_geometry}c), a per-voxel PCA with only $k=18$ components already saturates query accuracy (Fig.~\ref{fig:vl_geometry}d), so the features are locally low-rank and a compact code per vertex suffices.

\section{Evaluation}
\label{sec:eval}

We evaluate whether \methodname's joint training of multiple modalities in one octree improves geometry (Sec.~\ref{sec:eval:geom}) and radiance (Sec.~\ref{sec:eval:photo}). We also compare the compressed VL features to baselines in 3D and 2D queries (Sec.~\ref{sec:eval:vl}) and ablate the design choices in \methodname (Sec.~\ref{sec:eval:abl}).

\subsection{Experiment Setup}
\label{sec:eval:setup}

\noindent\textbf{Datasets.} We evaluate on Replica (8 scenes) \cite{replica19arxiv}, which provides semantic ground truth for open-vocabulary evaluation, and TUM (3 sequences) \cite{tum2012}, for which we generate pseudo-ground-truth labels using LatentAM's \cite{latentam2026} 10 classes and SAM~3 \cite{sam3_2026}.

\noindent\textbf{Baselines.} For geometry, we compare with nvblox \cite{millane_nvblox_2024} (TSDF fusion) and OREN \cite{oren26}. For radiance and VL, we compare with the offline LangSplatV2 \cite{langsplatv2_2025} and two online methods, OnlineLangSplat (OLS) \cite{onlinelangsplat2025} and LatentAM \cite{latentam2026}, running \methodname with a matched backbone for each since the baselines' backbones differ, CLIP \cite{clip2021} for LangSplatV2 and OLS, and TIPSv2-L \cite{tipsv2_2026} for LatentAM. Absolute scores vary with the backbone, but \methodname's advantage over its matched baseline holds consistently across both. All methods use the ground-truth camera poses. Since the baselines are designed for 2D queries only, for 3D queries we inverse-distance-weight the features within 15\,cm of the query point.

\noindent\textbf{Metrics.} For geometry, we report mesh accuracy, completion, Chamfer distance, and F-score at 5\,cm, plus the mean absolute error (MAE) of the SDF and its gradient near ($|s| \le 10$\,cm) and far from the surface. For appearance, we report PSNR, SSIM, LPIPS, and depth L1. For VL, we report mIoU, mean accuracy (mAcc), and mAP in 3D and 2D, plus feature cosine similarity for reconstruction fidelity. We measure efficiency by FPS and peak allocated GPU memory on an NVIDIA RTX 5090.

\noindent\textbf{Implementation.} Unless stated otherwise, we use the same hyperparameters for all scenes. We use an 8-level octree with $r_0=0.1$\,m, and encode radiance as a per-vertex $\bbR^3$ vector with sigmoid activation. The VL dictionary caps at $G_{\max}=256$ atoms, with growth controlled by energy threshold $\epsilon_{\mathrm{dict}}=0.002$ and decay $\rho=0.9$.

\vspace{-0.5em}
\subsection{Qualitative Results}
\label{sec:eval:qual}

\begin{figure*}[t]
    \centering
    \includegraphics[width=\linewidth]{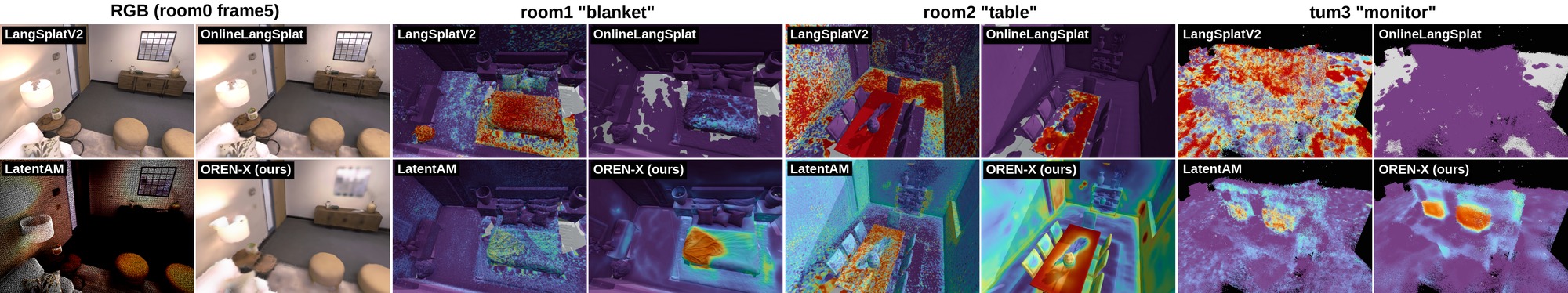}
    \caption{\textbf{Comparison on radiance and vision-language.} \methodname vs.\ LangSplatV2, OnlineLangSplat and LatentAM. Left: RGB rendering of Replica \texttt{room0}. Remaining columns: open-vocabulary query of \texttt{room1} (``blanket''), \texttt{room2} (``table'') and TUM 3 (``monitor'').} \label{fig:qual_compare}
    \vspace{-1.5em}
\end{figure*}

Fig.~\ref{fig:teaser} shows \methodname's modalities on a Replica scene, and Fig.~\ref{fig:qual_compare} compares our method with the radiance and VL baselines.
In Fig.~\ref{fig:qual_compare}, \methodname renders clean images, slightly smoother than LangSplatV2 and OLS, while LatentAM shows speckle artifacts.
For open-vocabulary queries, the baselines produce noisy, diffuse, or missing responses, whereas \methodname responds strongly, densely, and smoothly on the queried objects (blanket, table, monitor).

\vspace{-0.5em}
\subsection{Geometry}
\label{sec:eval:geom}

In Table~\ref{tab:geom}, nvblox has the best mesh, since TSDF fusion reproduces the visible surface directly, but its truncated field has the largest SDF and gradient errors.
Compared with OREN, \methodname reduces the near-surface SDF error by $33\%$ and the gradient error by $17\%$, with on-par far-field and mesh metrics while maintaining three more modalities, since occupancy and radiance supervise the SDF near the surface.

\vspace{-0.5em}
\subsection{Appearance}
\label{sec:eval:photo}

In Table~\ref{tab:photo}, LangSplatV2 has the best images on both datasets but the worst depth, since its offline-optimized Gaussians are not tied to a surface, and OLS shows the same trade-off to a lesser extent.
LatentAM, whose zero-degree spherical harmonics are similar to our RGB colors, has the best depth on Replica but the worst images.
\methodname renders along its SDF surface, balancing the two, with the second-best depth and better images than LatentAM on Replica, and clearly the best depth on the real TUM sequences.

\subsection{Open-Vocabulary Understanding}
\label{sec:eval:vl}

In Table~\ref{tab:vl}, one of the two \methodname backbones is best in 46 of the 48 Replica scene-metric entries.
On TUM, \methodname with TIPSv2-L remains best on all 3D query metrics and improves mIoU and mAcc by nearly $50\%$, while LatentAM and OLS lead in 2D since they are optimized specifically for 2D.
OLS reconstructs 2D features best in every Replica scene but trails in text queries. \methodname with CLIP mostly has the lowest fidelity but is best in most mIoU and mAcc entries. So, VL fidelity does not determine accuracy.

\begin{table}[t]
    \centering
    \caption{\textbf{Geometry} on Replica (8-scene mean). Mesh distances at a 5\,cm threshold, matching OREN's protocol. Best in \textbf{bold} and shaded green, second yellow.}
    \label{tab:geom}
    \resizebox{0.8\linewidth}{!}{\begin{tabular}{lccc}
        \toprule
        Metric & nvblox & OREN \cite{oren26} & \methodname (ours) \\
        \midrule
        Accuracy [cm] $\downarrow$ & \cellcolor{bestcell}\textbf{1.00} & 2.50 & \cellcolor{secondcell}2.49 \\
        Completion [cm] $\downarrow$ & 2.17 & \cellcolor{bestcell}\textbf{2.12} & \cellcolor{secondcell}2.14 \\
        Chamfer [cm] $\downarrow$ & \cellcolor{bestcell}\textbf{1.58} & 2.31 & \cellcolor{secondcell}2.31 \\
        F-score [\%] $\uparrow$ & \cellcolor{bestcell}\textbf{93.27} & \cellcolor{secondcell}90.73 & 90.68 \\
        \midrule
        SDF MAE, All [cm] $\downarrow$ & 4.44 & \cellcolor{secondcell}2.25 & \cellcolor{bestcell}\textbf{1.92} \\
        SDF MAE, Near [cm] $\downarrow$ & 5.47 & \cellcolor{secondcell}1.70 & \cellcolor{bestcell}\textbf{1.14} \\
        SDF MAE, Far [cm] $\downarrow$ & 3.73 & \cellcolor{bestcell}\textbf{2.58} & \cellcolor{secondcell}2.62 \\
        \midrule
        Grad. MAE, All [rad] $\downarrow$ & 0.272 & \cellcolor{secondcell}0.160 & \cellcolor{bestcell}\textbf{0.147} \\
        Grad. MAE, Near [rad] $\downarrow$ & 0.369 & \cellcolor{secondcell}0.122 & \cellcolor{bestcell}\textbf{0.101} \\
        Grad. MAE, Far [rad] $\downarrow$ & 0.222 & \cellcolor{bestcell}\textbf{0.176} & \cellcolor{secondcell}0.176 \\
        \bottomrule
    \end{tabular}}
\vspace{-1em}
\end{table}

\begin{table}[t]
    \centering
    \caption{\textbf{Photometric} quality on Replica and TUM (mean over each dataset). Best in \textbf{bold} and shaded green, second yellow.}
    \label{tab:photo}
    \resizebox{\linewidth}{!}{\begin{tabular}{lcccccccc}
        \toprule
        & \multicolumn{2}{c}{PSNR$\uparrow$} & \multicolumn{2}{c}{SSIM$\uparrow$} & \multicolumn{2}{c}{LPIPS$\downarrow$} & \multicolumn{2}{c}{Depth L1 (mm)$\downarrow$} \\
        \cmidrule(lr){2-3} \cmidrule(lr){4-5} \cmidrule(lr){6-7} \cmidrule(lr){8-9}
        Method & Replica & TUM & Replica & TUM & Replica & TUM & Replica & TUM \\
        \midrule
        LangSplatV2 \cite{langsplatv2_2025} & \cellcolor{bestcell}\textbf{36.00} & \cellcolor{bestcell}\textbf{19.74} & \cellcolor{bestcell}\textbf{0.96} & \cellcolor{bestcell}\textbf{0.78} & \cellcolor{bestcell}\textbf{0.0683} & \cellcolor{bestcell}\textbf{0.2281} & 38.50 & 77.72 \\
        OLS \cite{onlinelangsplat2025} & \cellcolor{secondcell}32.38 & \cellcolor{secondcell}15.61 & \cellcolor{secondcell}0.91 & \cellcolor{secondcell}0.62 & \cellcolor{secondcell}0.2359 & \cellcolor{secondcell}0.4377 & 17.52 & \cellcolor{secondcell}36.21 \\
        LatentAM \cite{latentam2026} & 23.77 & 12.84 & 0.76 & 0.49 & 0.3789 & 0.6057 & \cellcolor{bestcell}\textbf{6.29} & 54.98 \\
        \methodname (ours) & 28.01 & 14.21 & 0.88 & 0.56 & 0.2749 & 0.5038 & \cellcolor{secondcell}10.02 & \cellcolor{bestcell}\textbf{17.12} \\
        \bottomrule
    \end{tabular}}
\end{table}

\definecolor{bestcell}{RGB}{198,239,206}
\definecolor{secondcell}{RGB}{255,235,156}

\begin{table*}[t]
\centering
\scriptsize
\caption{\textbf{Open-vocabulary} understanding on Replica and TUM (mean). Best per column shaded green and bold, second yellow.}
\label{tab:vl}
\resizebox{\linewidth}{!}{\begin{tabular}{llcccccccccccccccccc}
\toprule
Method & Metric $\uparrow$ & \multicolumn{2}{c}{Room0} & \multicolumn{2}{c}{Room1} & \multicolumn{2}{c}{Room2} & \multicolumn{2}{c}{Office0} & \multicolumn{2}{c}{Office1} & \multicolumn{2}{c}{Office2} & \multicolumn{2}{c}{Office3} & \multicolumn{2}{c}{Office4} & \multicolumn{2}{c}{TUM (mean)} \\
\cmidrule(lr){3-4} \cmidrule(lr){5-6} \cmidrule(lr){7-8} \cmidrule(lr){9-10} \cmidrule(lr){11-12} \cmidrule(lr){13-14} \cmidrule(lr){15-16} \cmidrule(lr){17-18} \cmidrule(lr){19-20}
 & & 3D & 2D & 3D & 2D & 3D & 2D & 3D & 2D & 3D & 2D & 3D & 2D & 3D & 2D & 3D & 2D & 3D & 2D \\
\midrule
\multirow{4}{*}{\shortstack[l]{LangSplatV2\\[1pt](CLIP)}} & Cos-Sim & \cellcolor{secondcell}0.858 & \cellcolor{secondcell}0.918 & \cellcolor{bestcell}\textbf{0.882} & \cellcolor{secondcell}0.924 & \cellcolor{secondcell}0.865 & \cellcolor{secondcell}0.923 & \cellcolor{bestcell}\textbf{0.885} & \cellcolor{secondcell}0.917 & \cellcolor{secondcell}0.893 & \cellcolor{secondcell}0.929 & \cellcolor{secondcell}0.864 & \cellcolor{secondcell}0.914 & \cellcolor{secondcell}0.852 & \cellcolor{secondcell}0.904 & \cellcolor{secondcell}0.868 & \cellcolor{secondcell}0.921 & 0.783 & \cellcolor{bestcell}\textbf{0.903} \\
 & mIoU & 19.10 & 28.00 & 18.40 & 26.80 & 18.30 & 25.00 & 13.30 & 23.60 & 10.00 & 12.30 & 21.60 & 34.70 & 15.30 & 26.70 & 23.40 & \cellcolor{secondcell}29.80 & 9.57 & 42.10 \\
 & mAcc & 29.80 & 36.00 & 32.80 & 37.70 & 31.90 & 40.30 & 22.60 & 35.90 & 24.10 & 27.60 & 30.70 & 42.90 & 26.00 & 35.90 & 41.80 & 47.20 & 28.30 & \cellcolor{secondcell}61.77 \\
 & mAP & 24.10 & 32.70 & 26.80 & 36.00 & 25.30 & 31.40 & 25.70 & 31.20 & 19.20 & 27.70 & 26.60 & 32.70 & 21.80 & 29.50 & 31.70 & \cellcolor{bestcell}\textbf{42.00} & 16.83 & 50.13 \\
\midrule
\multirow{4}{*}{\shortstack[l]{OLS\\[1pt](CLIP)}} & Cos-Sim & 0.738 & \cellcolor{bestcell}\textbf{0.945} & 0.854 & \cellcolor{bestcell}\textbf{0.952} & 0.768 & \cellcolor{bestcell}\textbf{0.951} & 0.830 & \cellcolor{bestcell}\textbf{0.939} & 0.854 & \cellcolor{bestcell}\textbf{0.938} & 0.765 & \cellcolor{bestcell}\textbf{0.934} & 0.790 & \cellcolor{bestcell}\textbf{0.939} & 0.814 & \cellcolor{bestcell}\textbf{0.942} & \cellcolor{bestcell}\textbf{0.864} & \cellcolor{secondcell}0.901 \\
 & mIoU & 20.30 & 30.00 & 21.00 & 33.10 & 24.40 & \cellcolor{secondcell}30.90 & 13.70 & 24.60 & 9.90 & 9.80 & 20.70 & 28.80 & 15.30 & 23.40 & 19.60 & 21.20 & \cellcolor{secondcell}22.23 & 43.40 \\
 & mAcc & 30.50 & 38.40 & 35.20 & 45.40 & 36.50 & 40.70 & 26.10 & 37.80 & 18.50 & 18.40 & 28.60 & 34.50 & 22.50 & 29.80 & 29.90 & 31.10 & 34.27 & 53.10 \\
 & mAP & 37.40 & \cellcolor{bestcell}\textbf{40.10} & 39.00 & 41.70 & 41.00 & 40.10 & 24.10 & 26.30 & 18.40 & 20.60 & 29.90 & 34.30 & 28.60 & 30.80 & 38.90 & 35.50 & 32.77 & \cellcolor{bestcell}\textbf{55.30} \\
\midrule
\multirow{4}{*}{\shortstack[l]{LatentAM\\[1pt](TIPSv2-L)}} & Cos-Sim & 0.799 & 0.819 & 0.793 & 0.821 & 0.791 & 0.811 & 0.797 & 0.813 & 0.810 & 0.826 & 0.804 & 0.817 & 0.791 & 0.817 & 0.804 & 0.821 & 0.733 & 0.803 \\
 & mIoU & 20.00 & 28.60 & 18.20 & 31.20 & 18.40 & 21.80 & 19.30 & 23.70 & 14.90 & 21.80 & 16.20 & 26.20 & 14.70 & 22.90 & 22.50 & 23.40 & 16.93 & \cellcolor{bestcell}\textbf{50.77} \\
 & mAcc & 34.60 & 44.80 & 36.70 & 52.80 & 34.20 & 41.70 & 32.90 & 43.50 & 28.50 & 33.20 & 30.40 & 37.90 & 26.60 & 36.30 & 38.90 & 47.70 & 31.60 & \cellcolor{bestcell}\textbf{67.70} \\
 & mAP & 31.80 & 32.40 & 28.40 & 35.90 & 29.80 & 34.20 & 29.10 & 31.60 & 26.10 & 27.10 & 25.20 & 30.70 & 23.70 & 27.10 & 35.50 & 34.00 & 23.13 & \cellcolor{secondcell}51.33 \\
\midrule
\multirow{4}{*}{\shortstack[l]{OREN-X\\[1pt](TIPSv2-L)}} & Cos-Sim & \cellcolor{bestcell}\textbf{0.898} & 0.884 & \cellcolor{secondcell}0.880 & 0.886 & \cellcolor{bestcell}\textbf{0.888} & 0.883 & \cellcolor{secondcell}0.867 & 0.876 & \cellcolor{bestcell}\textbf{0.894} & 0.884 & \cellcolor{bestcell}\textbf{0.872} & 0.872 & \cellcolor{bestcell}\textbf{0.876} & 0.885 & \cellcolor{bestcell}\textbf{0.894} & 0.875 & \cellcolor{secondcell}0.850 & 0.846 \\
 & mIoU & \cellcolor{bestcell}\textbf{33.09} & \cellcolor{bestcell}\textbf{37.34} & \cellcolor{secondcell}28.27 & \cellcolor{bestcell}\textbf{34.81} & \cellcolor{secondcell}27.07 & 27.51 & \cellcolor{bestcell}\textbf{30.13} & \cellcolor{bestcell}\textbf{28.82} & \cellcolor{secondcell}23.41 & \cellcolor{secondcell}25.00 & \cellcolor{secondcell}32.43 & \cellcolor{secondcell}34.90 & \cellcolor{secondcell}26.38 & \cellcolor{secondcell}27.99 & \cellcolor{secondcell}27.24 & 28.21 & \cellcolor{bestcell}\textbf{33.21} & \cellcolor{secondcell}48.86 \\
 & mAcc & \cellcolor{secondcell}50.60 & \cellcolor{bestcell}\textbf{53.98} & \cellcolor{bestcell}\textbf{54.83} & \cellcolor{secondcell}61.22 & \cellcolor{secondcell}47.92 & \cellcolor{secondcell}47.81 & \cellcolor{bestcell}\textbf{48.60} & \cellcolor{secondcell}47.12 & \cellcolor{secondcell}37.94 & \cellcolor{secondcell}38.56 & \cellcolor{secondcell}56.70 & \cellcolor{secondcell}54.14 & \cellcolor{secondcell}43.92 & \cellcolor{secondcell}44.00 & \cellcolor{secondcell}50.17 & \cellcolor{secondcell}53.33 & \cellcolor{bestcell}\textbf{50.42} & 61.17 \\
 & mAP & \cellcolor{bestcell}\textbf{49.02} & \cellcolor{secondcell}39.81 & \cellcolor{bestcell}\textbf{50.83} & \cellcolor{secondcell}47.48 & \cellcolor{secondcell}44.91 & \cellcolor{secondcell}41.13 & \cellcolor{secondcell}38.17 & \cellcolor{bestcell}\textbf{33.75} & \cellcolor{bestcell}\textbf{34.90} & \cellcolor{bestcell}\textbf{31.91} & \cellcolor{bestcell}\textbf{46.50} & \cellcolor{bestcell}\textbf{38.16} & \cellcolor{secondcell}35.20 & \cellcolor{secondcell}31.52 & \cellcolor{secondcell}39.62 & 36.60 & \cellcolor{bestcell}\textbf{34.18} & 45.85 \\
\midrule
\multirow{4}{*}{\shortstack[l]{OREN-X\\[1pt](CLIP)}} & Cos-Sim & 0.790 & 0.776 & 0.776 & 0.776 & 0.773 & 0.766 & 0.753 & 0.756 & 0.763 & 0.757 & 0.765 & 0.762 & 0.761 & 0.769 & 0.766 & 0.753 & 0.761 & 0.756 \\
 & mIoU & \cellcolor{secondcell}27.80 & \cellcolor{secondcell}30.77 & \cellcolor{bestcell}\textbf{28.74} & \cellcolor{secondcell}34.04 & \cellcolor{bestcell}\textbf{31.09} & \cellcolor{bestcell}\textbf{32.69} & \cellcolor{secondcell}28.13 & \cellcolor{secondcell}26.60 & \cellcolor{bestcell}\textbf{25.00} & \cellcolor{bestcell}\textbf{28.43} & \cellcolor{bestcell}\textbf{39.37} & \cellcolor{bestcell}\textbf{40.66} & \cellcolor{bestcell}\textbf{29.35} & \cellcolor{bestcell}\textbf{29.03} & \cellcolor{bestcell}\textbf{38.93} & \cellcolor{bestcell}\textbf{42.22} & 19.42 & 34.21 \\
 & mAcc & \cellcolor{bestcell}\textbf{50.66} & \cellcolor{secondcell}51.79 & \cellcolor{secondcell}53.19 & \cellcolor{bestcell}\textbf{61.24} & \cellcolor{bestcell}\textbf{53.05} & \cellcolor{bestcell}\textbf{54.81} & \cellcolor{secondcell}46.92 & \cellcolor{bestcell}\textbf{47.85} & \cellcolor{bestcell}\textbf{44.99} & \cellcolor{bestcell}\textbf{44.11} & \cellcolor{bestcell}\textbf{61.72} & \cellcolor{bestcell}\textbf{57.89} & \cellcolor{bestcell}\textbf{49.70} & \cellcolor{bestcell}\textbf{46.34} & \cellcolor{bestcell}\textbf{62.48} & \cellcolor{bestcell}\textbf{66.72} & \cellcolor{secondcell}39.70 & 47.70 \\
 & mAP & \cellcolor{secondcell}42.85 & 36.67 & \cellcolor{secondcell}44.76 & \cellcolor{bestcell}\textbf{48.76} & \cellcolor{bestcell}\textbf{48.35} & \cellcolor{bestcell}\textbf{44.05} & \cellcolor{bestcell}\textbf{39.38} & \cellcolor{secondcell}32.87 & \cellcolor{secondcell}28.40 & \cellcolor{secondcell}29.61 & \cellcolor{secondcell}46.25 & \cellcolor{secondcell}35.07 & \cellcolor{bestcell}\textbf{35.28} & \cellcolor{bestcell}\textbf{34.18} & \cellcolor{bestcell}\textbf{43.20} & \cellcolor{secondcell}40.36 & \cellcolor{secondcell}34.12 & 42.85 \\
\bottomrule
\end{tabular}}
\vspace{-1.5em}
\end{table*}

\subsection{Ablation Studies}
\label{sec:eval:abl}

Unless stated otherwise, ablations use Replica \texttt{room0}.

\begin{table*}[t]
    \centering
    \caption{\textbf{Efficiency on Replica \texttt{room0}.} Throughput and peak GPU memory of \methodname as modalities are added, and of the baselines.}
    \label{tab:profiling}
    \scriptsize
    \setlength{\tabcolsep}{4pt}
    \resizebox{0.8\linewidth}{!}{\begin{tabular}{lccccccccc}
        \toprule
        & \multicolumn{5}{c}{\methodname} & \multicolumn{4}{c}{Baselines} \\
        \cmidrule(lr){2-6} \cmidrule(lr){7-10}
        Metric & SDF & SDF+OCC & SDF+Rad & SDF+OCC+Rad & SDF+OCC+Rad+VL & LatentAM & OnlineLangSplat & LangSplatV2 & nvblox \\
        \midrule
        FPS$\uparrow$ & 80.06 & 71.72 & 60.89 & 54.07 & 32.37 & 34.11 & 1.79 & 0.42 & 73.66 \\
        GPU Peak\,(GiB)$\downarrow$ & 0.28 & 0.34 & 0.78 & 0.87 & 1.01 & 8.66 & 11.32 & 13.18 & 1.00 \\
        \bottomrule
    \end{tabular}}
    \vspace{-2.5em}
\end{table*}

\noindent\textbf{Efficiency.}
Table~\ref{tab:profiling} reports throughput and peak GPU memory as modalities are added to \methodname.
Each added modality lowers throughput moderately and the VL modality costs the most, yet \methodname runs all four modalities in real time within about 1\,GiB of GPU memory, and with the SDF alone is faster than nvblox and uses less memory.
Among the VL methods, LatentAM reaches a similar throughput but needs about eight times more memory, while OnlineLangSplat and LangSplatV2 are one to two orders of magnitude slower.

\begin{figure}[t]
    \centering
    \includegraphics[width=0.9\linewidth]{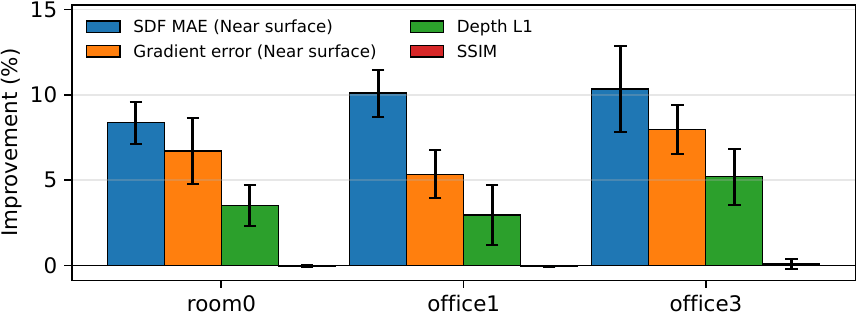}
    \caption{\textbf{Per-vertex bias correction vs.\ vanilla Adam.} Relative improvement from per-vertex bias correction on near-surface SDF MAE, near-surface gradient error, mean depth L1 and SSIM. Error bars are calculated over five seeds.}
    \label{fig:online_adam}
\vspace{-1.5em}
\end{figure}

\begin{figure}[t]
    \centering
    \includegraphics[width=\linewidth]{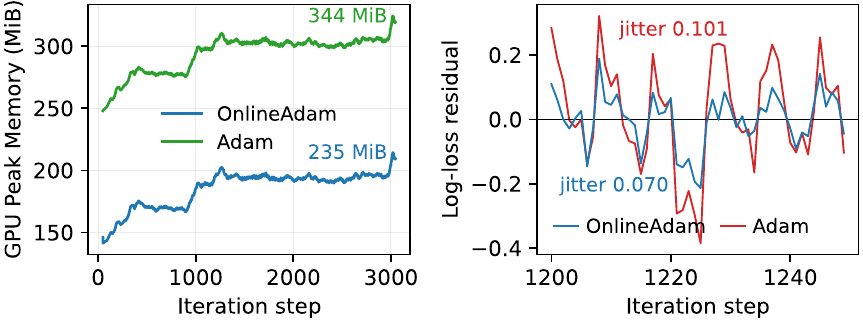}
    \caption{\textbf{Per-vertex bias correction vs.\ vanilla Adam: memory and stability.} Left: GPU peak memory over training with the SDF, OCC and radiance modalities on Replica \texttt{room0}. Right: detrended loss residual over a window. The legend reports each optimizer's jitter (the standard deviation of this residual over the whole run).}
    \label{fig:online_adam_mem_stability}
    \vspace{-1.5em}
\end{figure}

\noindent\textbf{Per-vertex bias correction.}
Compared with vanilla Adam, per-vertex bias correction consistently reduces the near-surface SDF, gradient and depth errors, while SSIM is unaffected (Fig.~\ref{fig:online_adam}), since the correction matters for rarely observed vertices near the surface, and updating moments only for active vertices lowers peak memory by about a third and reduces loss jitter (Fig.~\ref{fig:online_adam_mem_stability}).

\begin{figure}
    \centering
    \includegraphics[width=\linewidth]{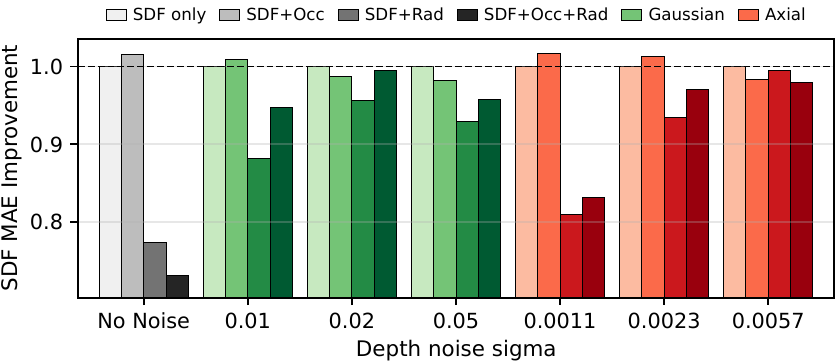}
    \caption{\textbf{Depth-noise robustness.} Replica \texttt{room0}. Near-surface SDF MAE of each modality
    combination, normalized by the SDF-only variant at the same noise level
    (lower is better, the dashed line is SDF-only). Two noise models are used:
    Gaussian noise $\calN(0, \sigma^2)$ and axial noise $\calN(0, (\sigma z)^2)$,
    where $z$ is the depth value.}
    \label{fig:noise}
\vspace{-1em}
\end{figure}

\noindent\textbf{Cross-modality synergy and depth noise.}
Table~\ref{tab:noise_robustness} and Fig.~\ref{fig:noise} add modalities under increasing depth noise.
Radiance reduces the SDF error most on clean depth, as photometric gradients pull the zero level set onto the surface but its benefit shrinks as noise grows.
Occupancy helps only under strong noise, where confident logits correct the SDF sign, and all modalities are best on clean depth and the strongest noise.

\begin{table}[t]
    \centering
    \caption{\textbf{Depth-noise robustness.} Replica \texttt{room0}. Near-surface SDF MAE (mm) across injected depth-noise levels, for Gaussian noise and axial noise. Best per column in \textbf{bold}, shaded green.}
    \label{tab:noise_robustness}
    \small
    \resizebox{0.9\linewidth}{!}{\begin{tabular}{lccccccc}
        \toprule
         &  & \multicolumn{3}{c}{Gaussian $\sigma$} & \multicolumn{3}{c}{Axial $\sigma$} \\
        \cmidrule(lr){3-5} \cmidrule(lr){6-8}
        Variant & No Noise & 0.01 & 0.02 & 0.05 & 0.0011 & 0.0023 & 0.0057 \\
        \midrule
        SDF & 15.59 & 13.67 & 13.55 & 20.07 & 14.66 & 13.88 & 24.32 \\
        SDF+OCC & 15.83 & 13.78 & 13.38 & 19.70 & 14.90 & 14.06 & 23.91 \\
        SDF+Rad & 12.06 & \cellcolor{bestcell}\textbf{12.04} & \cellcolor{bestcell}\textbf{12.95} & \cellcolor{bestcell}\textbf{18.66} & \cellcolor{bestcell}\textbf{11.87} & \cellcolor{bestcell}\textbf{12.97} & 24.19 \\
        SDF+OCC+Rad & \cellcolor{bestcell}\textbf{11.40} & 12.94 & 13.47 & 19.23 & 12.20 & 13.47 & \cellcolor{bestcell}\textbf{23.81} \\
        \bottomrule
    \end{tabular}}
\end{table}

\noindent\textbf{SDF-induced surface location.} Replacing the segment midpoint with the SDF zero crossing (Table~\ref{tab:surface_qstar}) leaves PSNR and SSIM unchanged but reduces LPIPS by $7\%$, indicating better perceptual quality.

\begin{table}[t]
    \centering
    \caption{\textbf{SDF-induced surface location.} Replica \texttt{room0}. Rendering with the sample placed at the SDF zero crossing vs.\ the segment midpoint. Best per column in \textbf{bold}, shaded green.}
    \label{tab:surface_qstar}
    \scriptsize
    \setlength{\tabcolsep}{4pt}
    \resizebox{0.7\linewidth}{!}{\begin{tabular}{lcccc}
        \toprule
        Variant & PSNR$\uparrow$ & SSIM$\uparrow$ & LPIPS$\downarrow$ & FPS$\uparrow$ \\
        \midrule
        SDF zero crossing & 28.35 & 0.87 & \cellcolor{bestcell}\textbf{0.2978} & 64.84 \\
        Segment midpoint & \cellcolor{bestcell}\textbf{28.36} & \cellcolor{bestcell}\textbf{0.88} & 0.3209 & \cellcolor{bestcell}\textbf{66.28} \\
        \bottomrule
    \end{tabular}}
\vspace{-0.75em}
\end{table}

\noindent\textbf{Radiance representation.} Spherical harmonics (Table~\ref{tab:abl_rast}) improve quality only marginally on the largely Lambertian Replica scenes, using $2.5\times$ memory, so we keep RGB.

\begin{table}[t]
    \centering
    \caption{\textbf{RGB vs.\ spherical-harmonic (SH) color.} Replica \texttt{room0}. Mem.\ is peak GPU memory per training step, and FPS is training throughput. Best per column in \textbf{bold}, shaded green.}
    \label{tab:abl_rast}
    \resizebox{0.9\linewidth}{!}{
    \begin{tabular}{lccccc}
        \toprule
        Repr. & PSNR$\uparrow$ & SSIM$\uparrow$ & LPIPS$\downarrow$ & Mem.\,(MiB)$\downarrow$ & FPS$\uparrow$ \\
        \midrule
        RGB & 28.32 & 0.8742 & 0.2981 & \cellcolor{bestcell}\textbf{242.8} & \cellcolor{bestcell}\textbf{32.75} \\
        SH1 & 28.30 & 0.8737 & 0.2953 & 313.8 & 32.07 \\
        SH2 & 28.40 & 0.8743 & 0.2947 & 433.5 & 31.92 \\
        SH3 & \cellcolor{bestcell}\textbf{28.46} & \cellcolor{bestcell}\textbf{0.8746} & \cellcolor{bestcell}\textbf{0.2942} & 608.4 & 31.45 \\
        \bottomrule
    \end{tabular}}
\vspace{-0.5em}
\end{table}

\noindent\textbf{Train/eval rasterizer.} In Table~\ref{tab:renderer_choice}, training with rays gives better quality with either evaluation backend and faster steps at more memory, while the tile-based rasterizer renders two orders of magnitude faster at a small quality cost, so the same octree serves both quality and speed.

\begin{table}[t]
    \centering
    \caption{\textbf{Train/eval rasterizer.} Replica \texttt{room0}. Ray- vs.\ tile-based rasterizer for training and evaluation. Train FPS and Mem.\ (peak GPU memory) depend only on the training rasterizer, and Eval FPS is measured at $1200\times680$. Best per column in \textbf{bold}.}
    \label{tab:renderer_choice}
    \setlength{\tabcolsep}{4pt}
    \resizebox{0.9\linewidth}{!}{\begin{tabular}{lcccccc}
        \toprule
        Train/Eval & PSNR$\uparrow$ & SSIM$\uparrow$ & LPIPS$\downarrow$ & Train FPS$\uparrow$ & Eval FPS$\uparrow$ & Mem.\,(MiB)$\downarrow$ \\
        \midrule
        ray/ray & \cellcolor{bestcell}\textbf{28.32} & 0.8742 & \cellcolor{bestcell}\textbf{0.2981} & \multirow{2}{*}{\textbf{31.80}} & 12.0 & \multirow{2}{*}{341.1} \\
        ray/tile & 27.68 & 0.8703 & 0.3036 &  & \cellcolor{bestcell}\textbf{1513.8} &  \\
        \midrule
        tile/ray & 26.77 & \cellcolor{bestcell}\textbf{0.8751} & 0.3154 & \multirow{2}{*}{29.25} & 12.0 & \multirow{2}{*}{\textbf{262.5}} \\
        tile/tile & 27.00 & \cellcolor{bestcell}\textbf{0.8751} & 0.3138 &  & 1456.7 &  \\
        \bottomrule
    \end{tabular}}
\vspace{-1em}
\end{table}

\noindent\textbf{Online dictionary.} In Table~\ref{tab:abl_codebook}, the online dictionary gives the best mAP at about a quarter of the memory of full features, since discarding low-variance directions removes noise from the cosine score.
An implicit MLP stores the VL field in even less memory but loses substantial accuracy, and its features must be decoded before querying.
An offline PCA dictionary degrades when centered, because the codes omit the feature mean, which dominates VL features (Fig.~\ref{fig:vl_geometry}a).
Accuracy saturates at $G_{\max}{=}256$, as the energy criterion stops adding atoms before reaching capacity, so a larger $G_{\max}$ only widens the stored coefficient vectors, which is why memory keeps growing. Sparse top-$K$ codes with fewer atoms keep most accuracy, using less memory.

\begin{table}[t]
    \centering
    \caption{\textbf{Online dictionary.} Replica \texttt{room0}. VL field stored as full per-vertex features, as implicit features decoded by an MLP, as codes over an offline PCA dictionary (with/without centering), and as codes from our online dictionary, with dense coefficients for varying $G_{\max}$ or sparse top-$K$ coefficients at $G_{\max}{=}256$. $|D|$ is the number of atoms the growth criterion activated. It equals $G_{\max}$ for the fixed-size offline dictionaries. Dense rows pad each coefficient vector to $G_{\max}$. Top-$K$ rows store $K$ coefficients and their indices. Mem.\ is the storage of the VL field (MB). Best per column in \textbf{bold}, shaded green.}
    \label{tab:abl_codebook}
    \setlength{\tabcolsep}{4pt}
    \resizebox{0.9\linewidth}{!}{
    \begin{tabular}{lcccccc}
        \toprule
        Variant & $G_{\max}$ & $|D|$ & Cosine$\uparrow$ & mAP$\uparrow$ & Mem.$\downarrow$ & FPS$\uparrow$ \\
        \midrule
        Full features & -- & -- & \cellcolor{bestcell}\textbf{0.907} & 45.87\% & 256.2 & 31.00 \\
        Implicit MLP & -- & -- & 0.902 & 41.59\% & 9.5 & 31.06 \\
        \midrule
        Offline (uncentered) & 256 & 256 & 0.905 & 46.66\% & 69.3 & \cellcolor{bestcell}\textbf{32.37} \\
        Offline (centered) & 256 & 256 & 0.736 & 44.54\% & 69.3 & \cellcolor{bestcell}\textbf{32.37} \\
        \midrule
        Online, dense coeff. & 64 & 64 & 0.845 & 40.96\% & 20.5 & -- \\
        Online, dense coeff. & 128 & 128 & 0.884 & 45.38\% & 36.8 & -- \\
        Online, dense coeff. & 256 & 228 & 0.898 & \cellcolor{bestcell}\textbf{49.02\%} & 69.3 & \cellcolor{bestcell}\textbf{32.37} \\
        Online, dense coeff. & 512 & 228 & 0.898 & \cellcolor{bestcell}\textbf{49.02\%} & 134.3 & -- \\
        Online, dense coeff. & 1024 & 228 & 0.898 & \cellcolor{bestcell}\textbf{49.02\%} & 264.3 & -- \\
        \midrule
        Online, top-$K{=}4$ & 256 & 228 & 0.795 & 32.74\% & \cellcolor{bestcell}\textbf{6.8} & 29.99 \\
        Online, top-$K{=}48$ & 256 & 228 & 0.878 & 45.38\% & 23.3 & 29.99 \\
        \bottomrule
    \end{tabular}}
\end{table}

\begin{table}[t]
    \centering
    \caption{\textbf{VL backbone comparison.} Replica (8-scene mean). VL reconstruction and open-vocabulary query quality of \methodname with four VL backbones, each cell 3D/2D. Best per 3D or 2D half in \textbf{bold}, shaded green, second yellow.}
    \label{tab:backbone}
    \setlength{\tabcolsep}{4pt}
    \resizebox{\linewidth}{!}{\begin{tabular}{lcccc}
        \toprule
        Backbone & Cos-Sim (3D/2D)$\uparrow$ & mIoU (3D/2D)$\uparrow$ & mAcc (3D/2D)$\uparrow$ & mAP (3D/2D)$\uparrow$ \\
        \midrule
        CLIP ViT-B/16 & 0.768/\colorbox{secondcell}{0.764} & \colorbox{secondcell}{31.05}/\colorbox{secondcell}{33.06} & \colorbox{bestcell}{\textbf{52.84}}/\colorbox{bestcell}{\textbf{53.84}} & 41.06/\colorbox{secondcell}{37.69} \\
        DINOv3.txt & 0.684/0.694 & 27.96/30.59 & 43.72/45.66 & 39.40/35.25 \\
        Talk2DINOv3-ViT-L & \colorbox{secondcell}{0.771}/0.748 & \colorbox{bestcell}{\textbf{33.50}}/\colorbox{bestcell}{\textbf{35.43}} & 46.87/48.01 & \colorbox{bestcell}{\textbf{42.42}}/\colorbox{bestcell}{\textbf{40.46}} \\
        TIPSv2-L & \colorbox{bestcell}{\textbf{0.884}}/\colorbox{bestcell}{\textbf{0.881}} & 28.50/30.57 & \colorbox{secondcell}{48.83}/\colorbox{secondcell}{50.02} & \colorbox{secondcell}{42.40}/37.54 \\
        \bottomrule
    \end{tabular}}
\end{table}

\noindent\textbf{VL backbone.} Table~\ref{tab:backbone} compares four VL backbones on Replica.
\methodname reconstructs TIPSv2-L features best and DINOv3.txt features worst, consistent with Fig.~\ref{fig:vl_geometry}a, where TIPSv2-L features concentrate most tightly around their mean and DINOv3.txt features spread the most. CLIP gives the best mAcc and Talk2DINOv3-ViT-L the best mIoU and mAP. Reconstruction fidelity again does not determine query accuracy, and the same dictionary hyperparameters work well across all four backbones without tuning.

\section{Conclusion}
\label{sec:conclusion}

We presented \methodname, an online multi-modal mapping framework that maintains SDF, occupancy, radiance and VL fields in a single octree with a common storage representation, shared indexing and joint training.
Joint training improves SDF accuracy, shared indexing enables real-time rendering and both 2D and 3D open-vocabulary queries, and an online dictionary compresses the VL field while improving query accuracy.
On Replica, \methodname maps in real time and outperforms prior methods in SDF accuracy and open-vocabulary segmentation.
On the real-world TUM RGB-D sequences \cite{tum2012}, it achieves the lowest depth error and the best 3D open-vocabulary segmentation.

\balance
{\small
\bibliographystyle{IEEEtran}
\bibliography{bib/main}
}

\end{document}